\RequirePackage[T1]{fontenc}
\documentclass[conference,letterpaper]{IEEEtran}
\IEEEoverridecommandlockouts
\usepackage{cite,amsmath,amssymb,graphicx,booktabs,array,url,stfloats,overpic,xcolor}
\usepackage[hidelinks]{hyperref}
\hypersetup{pdftitle={Can an Agent Rethink What It Knows? Selective Scene Memory Updates in a Changing World},pdfauthor={Zhekai Ren and Jiaming Wu}}
\newcommand{\figwide}[4]{\begin{figure*}[!tb]\centering\includegraphics[width=\textwidth]{figures/#1.pdf}\caption{#2}\label{#3}\end{figure*}}

\title{Can an Agent Rethink What It Knows?\\[0.3em]
{\large\textit{Selective Scene Memory Updates in a Changing World}}}

\author{
\IEEEauthorblockN{Zhekai Ren\textsuperscript{1,\textdagger}, Jiaming Wu\textsuperscript{1,2,\textdagger,*}}
\IEEEauthorblockA{\textsuperscript{1}Sensteed AI Lab\\
\textsuperscript{2}University College London\\
\textsuperscript{*}Corresponding author: \texttt{jiaming.wu.20@ucl.ac.uk}}
\thanks{\textsuperscript{\textdagger}These authors contributed equally to this work.}
}
\begin{document}
\maketitle
\pagestyle{plain}
\raggedbottom

\begin{abstract}
Task configurations in factories, warehouses, and everyday physical environments change over time. Physical AI systems therefore need to incorporate new facts into scene memory while preserving valid information about equipment, objects, and operating constraints. We study updates to this parametric semantic-memory layer and present a joint evaluation of correction and preservation under finite update budgets. At fixed current fact count and memory parameter count, the protocol combines independent single-fact updates, budget trajectories, old-to-new answer diagnostics, and first stable success. Using a frozen 4B language model with late-layer LoRA, we evaluate seven preparation conditions and six update budgets on four-field synthetic semantic configurations. In two exposure-matched history conditions, increasing the budget from eight to 32 updates raises joint success from 6.25\% to 94.79\% and 91.15\%, respectively, while exact obsolete-value outputs fall from 174/192 and 166/192 to zero. Constant32, which matches preparation steps but repeatedly trains on unchanged content during the first 32 events, achieves higher early joint success at eight steps (53.65\%); at 32 steps, it reaches 85.42\%, below both changing-history conditions. These results distinguish rapid correction under a small budget from simultaneous correctness of the new fact and the other three facts after further training. For both changing histories, most cases remain jointly correct at all subsequent measured points from step 16 onward. Output analysis shows that short-budget failures primarily retain the old value; repeated training on the new fact together with the other current facts progressively replaces these obsolete answers.

\end{abstract}
\begin{IEEEkeywords}
Physical AI, scene memory, parametric memory, memory updates, correction and preservation, update budget

\end{IEEEkeywords}

\section{Introduction}
Production changeovers, warehouse reorganization, and on-site collaboration alter the knowledge on which tasks depend. During a production changeover, for example, a workpiece-handling configuration may change while equipment conditions and other operating rules remain valid. In a warehouse, an object's assigned location may change while other objects' attributes and handling requirements must be retained. Physical AI systems that perceive, reason, or act in physical environments need to maintain current knowledge relevant to these on-site tasks.

We define \emph{scene memory} as a set of semantic facts associated with a particular site, device, or task configuration and revisable as that setting changes. It supplies currently valid information for queries and high-level task decisions. Perception, event reports, or human corrections provide update content; parametric memory writes that content into persistent state, from which subsequent queries retrieve the latest facts. Updating scene memory requires a local change to take effect while the remaining current facts stay correct.

We study \emph{finite-budget updates at fixed current fact count and memory parameter count}. Each evaluation independently starts from a prepared checkpoint, changes one fact, and preserves the others. Correction, preservation, joint success, and obsolete-value outputs are measured together at six update budgets. Our experiments use four-field synthetic semantic configurations and independent language queries to test the memory layer's outputs. Figure~\ref{fig:protocol} shows how the update histories are constructed. At recall time, the model receives its checkpoint and the current query; the facts are supplied by the parameterized state.

This paper examines the criteria for a successful memory update and its stability during subsequent training. An effective update must incorporate the new fact, preserve the remaining facts, and maintain joint correctness in later evaluations. At step 16, the two histories yield 153 and 144 cases with all four facts correct, respectively; in each condition, three of these cases become incorrect again at a later checkpoint. We therefore evaluate correctness at individual training checkpoints and identify the first checkpoint from which four-fact joint correctness persists at all subsequent measured points. This distinguishes checkpoint-level update success from sustained correctness.

We present a joint evaluation protocol at fixed memory size, measuring single-fact correction and preservation of the remaining facts within the same probe. We track the transition from old-value to new-value outputs, the model's relative preference between the two answers, and the onset of four-fact joint correctness that persists at subsequent measured checkpoints. We also report complete results across seven conditions, paired comparisons, and checkpoint-reload verification. The evaluation thus addresses three questions: how many training updates a correction requires, whether other facts remain correct, and whether joint correctness persists with continued training. Here, stability means correctness at every subsequent measured checkpoint through step 32 within the same correction run.

\section{Related Work}
LoRA represents adaptation through low-rank weight updates~\cite{lora}. We instantiate our evaluation with late-layer LoRA, freezing its learned input factor during rewriting. This factorization represents scene-level semantic state and fixes the writable parameter set for comparing update trajectories.

Model editing addresses targeted knowledge changes together with preservation of unrelated information. WISE uses side memory and routing~\cite{wise}; MEMOIR restricts updates through sparse, activation-dependent masks~\cite{memoir}. Editable Neural Networks and MEND train models or update rules for editability~\cite{editable,mend}. Our evaluation focuses on the budget axis: it measures how correction and preservation develop together as an update proceeds, complementing the architectural and optimization approaches in these studies.

Continual-learning research distinguishes acquisition of new information from retention of old information~\cite{plasticity}. We instantiate this distinction as updates to a semantic state with a fixed number of fields: one field receives a new value while the others retain their currently valid values. Joint correctness captures consistency of the complete state; output decomposition and stable-success budgets further characterize how that state is established.

Semantic knowledge in Physical AI must be connected to a concrete environment. SayCan combines language-model task knowledge with value functions for executable skills~\cite{saycan}; DynaMem maintains changing environment information using dynamic three-dimensional spatio-semantic memory~\cite{dynamem}. These studies demonstrate the role of scene information in physical tasks. We focus on the update interface of parametric semantic memory, evaluating whether a locally changed fact and the remaining state stay jointly correct. This perspective complements environment perception, spatial representation, and task execution.

\section{Joint Evaluation of Scene-Memory Updates}
\subsection{Scene State and Local Updates}
We represent each scene configuration as a set of currently valid semantic facts. An update replaces one field's old value with a new value while retaining the other fields. After the update, queries with different wording test each field, recording correctness of the new value, correctness of the remaining fields, and joint correctness of the complete configuration. This interface corresponds to local configuration revisions in factories, changes to object information in warehouses, and task-rule updates in everyday environments.

We implement this interface using four-field synthetic configurations. Teaching statements write field values into memory, checkpoints store the parameterized state, and evaluation questions test retrieval after writing. All configurations use the same number of fields and parameters, allowing comparisons of update budgets, preparation histories, and post-update correctness at fixed state capacity.
\subsection{Model and Controlled Semantic Configurations}
We freeze Qwen3-4B-Instruct-2507~\cite{qwen} and adapt query/value projections at layers 26 and 32. With $u$ indexing a configuration, each projection has the form
\begin{equation}
 h\mapsto W_0h+(\alpha/r)B_uA_uh,
\end{equation}
with rank $r=70$ and scale $\alpha=140$. Initial fitting updates A and B; subsequent histories and probes freeze the learned, configuration-specific A and update B. Each memory state contains 1,433,600 A+B parameters, of which 716,800 are updated during rewriting. Parameter count and fact count remain fixed across conditions. We express the probe budget as the number of successful optimization steps.

Four synthetic configurations, 024/063/175/190, use the discrete semantic fields drink, fruit, color, and music, each with four possible values. Initialization seeds are 51/52 and editing seeds 401/402, giving 16 paired configurations. These four configurations form the controlled test set for the memory layer. Eight initializations fit A+B for 400 steps. Teaching uses four templates per fact. Evaluation uses one canonical query and three unseen wordings per fact, held separate from teaching.

Editing uses FP32 master parameters and BF16 deployment, learning rate 0.0003, weight decay 0.01, gradient clipping 1, and target cross-entropy plus anchor KL with weight 1. TF32 is disabled. Each history event starts a fresh Adam optimizer; each probe keeps one optimizer for its 32 steps. The paired conditions use identical probe seeds and teaching schedules.

\subsection{Histories and Independent Correction Probes}

\begin{figure*}[!tb]
\centering
\begin{overpic}[width=\textwidth]{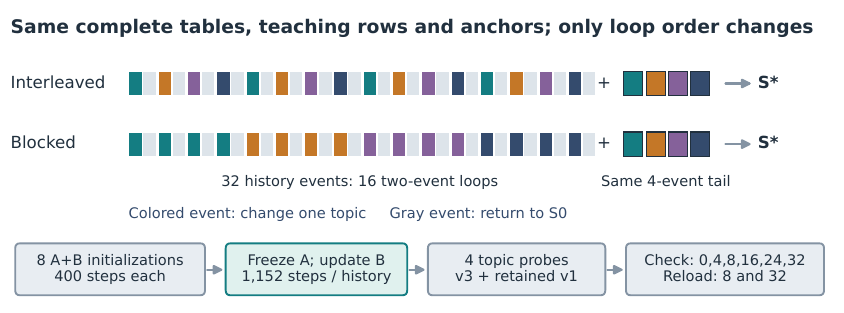}
\color[HTML]{647687}\linethickness{0.5pt}
\put(15.18356,25.22405){\framebox(13.84941,3.86404){}}
\put(29.08050,25.22405){\framebox(13.84941,3.86404){}}
\put(42.97743,25.22405){\framebox(13.84941,3.86404){}}
\put(56.87436,25.22405){\framebox(13.84941,3.86404){}}
\put(15.18356,18.03419){\framebox(13.84941,3.86404){}}
\put(29.08050,18.03419){\framebox(13.84941,3.86404){}}
\put(42.97743,18.03419){\framebox(13.84941,3.86404){}}
\put(56.87436,18.03419){\framebox(13.84941,3.86404){}}
\end{overpic}
\caption{Exposure-matched order intervention. Each colored event changes one topic; the following gray event returns to the initial table. Both sequences contain the same 16 two-event loops and common tail. Colors denote topic roles, which rotate across configurations. The final state forks into independent probes, each changing one topic to v3. The schematic represents the two primary histories; reference branches are specified in the text.}\label{fig:protocol}
\end{figure*}

A two-event loop changes one topic from $S_0$ and returns to $S_0$. Both histories contain four loops per topic. Interleaved uses topic order $[0,1,2,3]\times4$; blocked uses $[0,0,0,0,1,1,1,1,\ldots]$. Topic labels rotate across configurations. Seeds keyed to the complete fact table and its occurrence match the multisets of tables, teaching rows, and anchors. Both histories end at $S_0$ and share a four-event tail to $S^*$, containing v1 in each topic (Fig.~\ref{fig:protocol}). Each history uses 1,152 updates.

References comprise a short 128-step tail, constant32 (constant $S_0$ for 32 events plus the tail), and direct refits from the paired initial state at 32/128/512 steps. Constant32 holds the initial content constant while matching the primary histories' update count. Each of seven endpoints independently forks four probes, replacing one topic with reserved v3 and retaining the other three v1 values. The reserved value belongs to the predefined vocabulary and is newly introduced in that configuration's topic history. Each four-step block teaches one row per topic: eight updates provide two changed-fact rows; 32 provide eight. Training replays the complete current fact table; recall supplies only the current query.

\begin{table}[!tb]
\centering\small
\caption{Endpoint preparation after the common 400-step initialization. Probe budgets are additional. The two primary histories match complete exposure; the other branches are descriptive references.}
\label{tab:conditions}
\begin{tabular}{lrl}
\toprule
Branch & Steps & Preparation\\
\midrule
Interleaved & 1,152 & Alternating topic loops + tail\\
Blocked & 1,152 & Grouped topic loops + tail\\
Constant32 & 1,152 & Constant table + tail\\
Short & 128 & Tail only\\
Refit32 & 32 & Direct fit of current table\\
Refit128 & 128 & Direct fit of current table\\
Refit512 & 512 & Direct fit of current table\\
\bottomrule
\end{tabular}
\end{table}

\subsection{Joint Outcomes and Fixed Budgets}

For a four-fact probe that changes topic $p$, let $e_{i,s}(t)$ indicate exact correctness of fact $s$ at update budget $t$. Index $i$ includes configuration, initialization, edit seed, changed topic, and unseen wording. Joint correctness is
\begin{equation}
 J_i(t)=\prod_{s=1}^{4}e_{i,s}(t).
 \label{eq:joint}
\end{equation}
We report changed-fact accuracy, unchanged-fact accuracy, joint accuracy, and exact output of the superseded value. Equation~\eqref{eq:joint} requires the corrected fact and all three retained facts to be simultaneously correct. Recall uses the full stripped generated sentence. Changed-fact, unchanged-fact, and joint correctness have denominators 192, 576, and 192, respectively, at each probe budget. The observations are grouped by the four configurations and their paired seed configurations.

Budgets are evaluated on the fixed grid $\mathcal T=\{0,4,8,16,24,32\}$. We also record the first \emph{stable measured} success:
\begin{equation}
 \tau_i=\min\{t\in\mathcal T: J_i(t')=1\ \forall t'\in\mathcal T,t'\geq t\}.
 \label{eq:stable}
\end{equation}
An empty set is right-censored at 32. The resulting statistic summarizes the first successful checkpoint whose later measured checkpoints all remain correct, providing a common budget scale for comparing trajectories.

\subsection{Prespecified History-Order Contrast}

The primary contrast averages joint correctness over the paired configuration structure:
\begin{equation}
 \Delta_8=\mathbb E_u\mathbb E_{a,b,p,k}
 [J^{\rm blocked}_{u,a,b,p,k}(8)-J^{\rm interleaved}_{u,a,b,p,k}(8)].
\end{equation}
The prespecified screen requires $|\Delta_8|\geq0.20$, matching direction in at least three of four configuration means and both initialization means, plus comparability: both groups' current and final-probe fact accuracies $\geq95\%$, with gaps $\leq5$ points. Each branch contains 192 joint observations grouped within the 16 paired configurations.

\section{Results}
\subsection{Correction and Preservation Across Budgets}

At eight updates, changed-fact correctness is 12/192 and 14/192, while unchanged correctness remains 566/576 and 546/576. At 32 updates, changed-fact correctness rises to 190/192 and 188/192; joint success reaches 182/192 and 175/192. The three metrics in Fig.~\ref{fig:curves} distinguish learning the correction from retaining the remaining facts.

Constant32 reaches 103/192 (53.65\%) joint successes at eight steps and 164/192 (85.42\%) at 32. Its early lead is followed by a 32-step advantage of 9.38 and 5.73 percentage points for the two primary histories. This crossing distinguishes rapid initial correction from successful correction with preservation at a larger budget. The direct-refit references offer a further comparison: after 512 preparation steps, Refit512 reaches 102/192 at eight probe steps and 151/192 at 32. Table~\ref{tab:conditions} specifies each branch's exposure and preparation budget, and Appendix~\ref{app:grid} presents every branch at every predefined probe budget.

\begin{table*}[!tb]
\centering\small
\caption{Primary histories and descriptive references. Current facts: /192. Probe joint/new/old: /192; unchanged facts: /576; final facts: /768. References differ in exposure and/or preparation cost.}
\label{tab:rewrite}
\begin{tabular}{lrrrrrrr}
\toprule
Branch & Current facts & Joint@8 & New@8 & Unchanged@8 & Old@8 & Joint@32 & Facts@32\\
\midrule
Interleaved & 190/192 & 12/192 & 12/192 & 566/576 & 174/192 & 182/192 & 752/768\\
Blocked & 180/192 & 12/192 & 14/192 & 546/576 & 166/192 & 175/192 & 740/768\\
Short & 182/192 & 102/192 & 117/192 & 525/576 & 52/192 & 162/192 & 718/768\\
Constant32 & 185/192 & 103/192 & 117/192 & 545/576 & 60/192 & 164/192 & 722/768\\
Refit32 & 174/192 & 83/192 & 104/192 & 524/576 & 66/192 & 149/192 & 713/768\\
Refit128 & 180/192 & 89/192 & 105/192 & 535/576 & 70/192 & 151/192 & 715/768\\
Refit512 & 183/192 & 102/192 & 117/192 & 545/576 & 61/192 & 151/192 & 712/768\\
\bottomrule
\end{tabular}
\end{table*}

\figwide{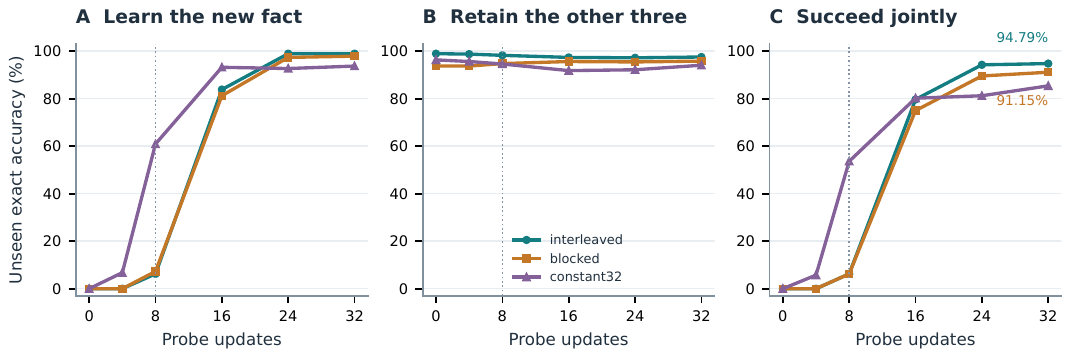}{Correction, preservation, and joint success at all predefined probe budgets. The vertical marker identifies the eight-step primary endpoint. The 0--100\% axes provide a common scale. Constant32 maintains constant initial content with the same preparation update count as the primary histories. Lines connect measured checkpoints.}{fig:curves}{}

\subsection{Tracking the Switch from Old to New Values}
At eight steps, old v1 answers account for 174/192 and 166/192 changed-topic outputs, while unchanged-fact correctness remains 566/576 and 546/576. At 32, both old-value counts reach zero (Fig.~\ref{fig:diagnostic}A). The output decomposition localizes the main early error to the target answer and directly tracks its replacement as the update proceeds.

Teacher-forced diagnostics score the complete new and old answers, including EOS, by summing token log probabilities. At eight steps, median margins (new minus old) are $-2.881$ and $-2.628$, versus $+1.367$ for Constant32. At zero steps, Constant32 starts with a more negative margin ($-11.371$) than Interleaved ($-9.596$), yet reaches a positive median earlier. The change in this ranking shows the value of following the likelihood trajectory in addition to recording the starting margin (Fig.~\ref{fig:diagnostic}B). The full-answer score includes answer length as part of its definition.

\figwide{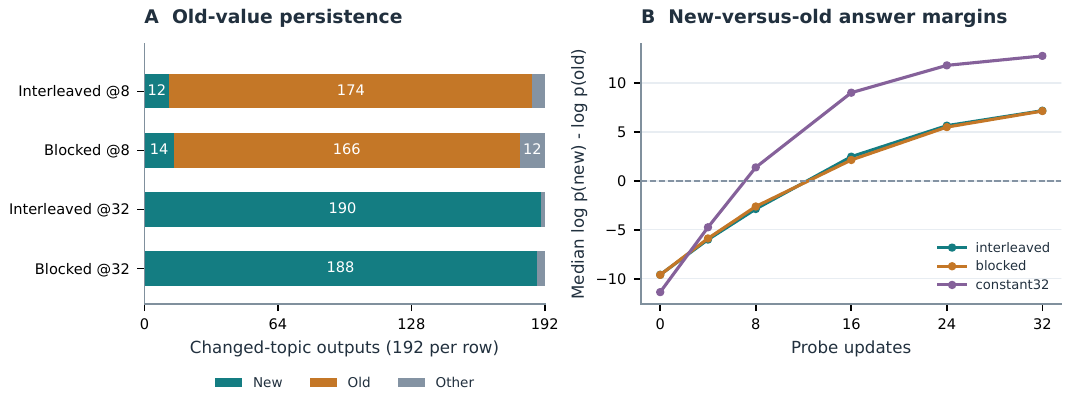}{Output and likelihood diagnostics. (A) Exact new, exact old, and all other changed-topic outputs; each row totals 192. (B) Median full-answer log-probability margins at measured checkpoints. Positive values favor the new answer; negative values favor the old answer. Constant32 starts with a more negative median and reaches a positive median earlier, showing a change in the ranking across budgets.}{fig:diagnostic}{}

\subsection{Stable Success on the Measured Grid}
Figure~\ref{fig:stable} summarizes Eq.~\eqref{eq:stable}. Interleaved first becomes stably correct at steps 8/16/24/32 for 12/138/30/2 cases; Blocked for 12/129/30/4. Ten and seventeen cases, respectively, are right-censored at 32. The largest group reaches stable success at 16 steps: 138/192 and 129/192 cases. Relative to the 12 cases already stable at eight steps, the next measured checkpoint therefore captures the main transition in both histories. This summary translates the joint-accuracy curves into the distribution of budgets at which successful corrections persist through the remaining measured checkpoints.

\figwide{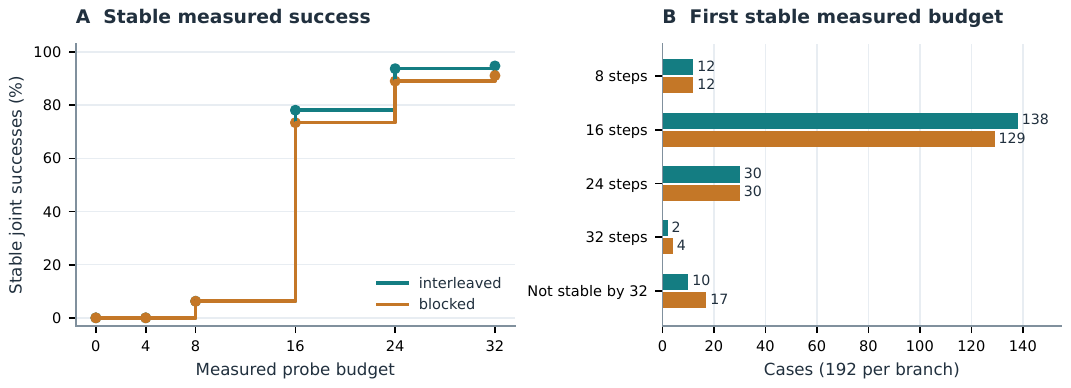}{Stable joint success on the fixed checkpoint grid. (A) Cumulative fraction reaching stable measured success by each budget. (B) First-success counts and cases right-censored at 32. Stability is defined as success at the first qualifying checkpoint and every later measured checkpoint through step 32.}{fig:stable}{}

\subsection{Paired History-Order Results}
Both histories achieve 12/192 joint successes at eight steps, giving $\Delta_8=0$. Configuration differences are $-4.17$, $+2.08$, $+2.08$, and $0$ percentage points; initialization differences are $-3.125$ and $+3.125$; and editing-seed differences are $-6.25$ and $+6.25$ (Fig.~\ref{fig:paired}). The paired presentation makes the opposite seed-level directions visible alongside their aggregate mean.

Current-fact scores are 190/192 (98.96\%) and 180/192 (93.75\%), a gap of 5.21 points against the protocol's 5-point comparability threshold. Final-probe fact scores are 752/768 and 740/768. We report these starting-state and final-state scores together with the fixed eight-step contrast, maintaining the same evaluation target across all paired configurations.

\figwide{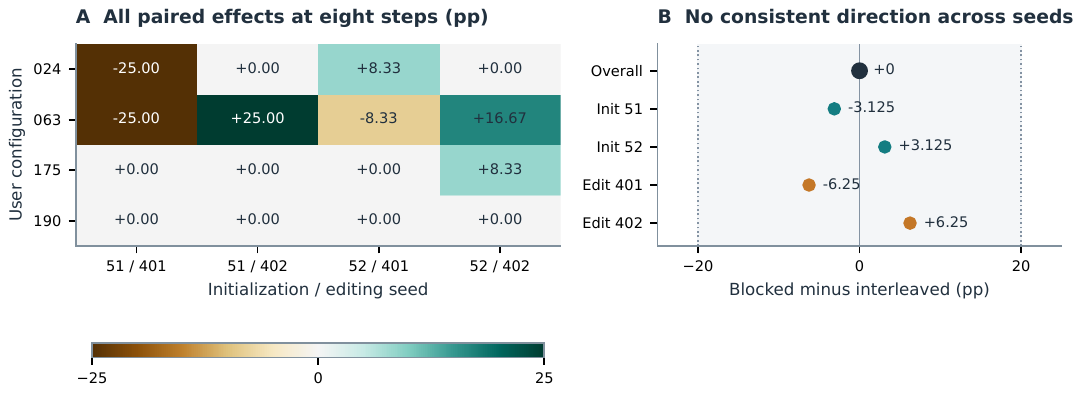}{Paired history-order results. (A) All 16 configuration/initialization/edit-seed contrasts, each averaging four topics and three unseen wordings. (B) Overall and seed-group means. Dotted lines mark the prespecified effect-size screening thresholds. Opposite seed-level directions average to zero.}{fig:paired}{}

\section{Discussion}
\subsection{Local Changes and Complete-State Consistency}
In factory or warehouse tasks, the value of a configuration update depends on its compatibility with the remaining current knowledge. During a production changeover, for instance, the handling configuration for a new workpiece must take effect while equipment conditions and other operating requirements remain valid. Joint correctness expresses this requirement as a check of the complete state after a single update.

In our experiments, changed-fact correctness reaches 190/192 and 188/192 at 32 steps, while joint correctness is 182/192 and 175/192. These differences distinguish acquisition of the new value from correctness of the complete configuration. Reporting correction, preservation, and joint-success curves together reveals how local updates and overall consistency evolve during training.

\subsection{Update Budgets and the Emergence of a Correct State}
On-site knowledge changes at different rates, making an explicit resource scale useful for memory updates. We measure semantic-memory updates in optimization steps and account for preparation history separately from probe budgets. At eight steps, Constant32 has 91 more joint successes than either primary history; at 32 steps, Interleaved and Blocked have 18 and 11 more joint successes than Constant32. These rankings describe the quality of the resulting state at different update budgets.

Output decomposition further characterizes the update process: old-value outputs dominate in both primary histories at eight steps, and exact obsolete-value outputs fall to zero by 32 steps. Likelihood margins track changing preferences between competing answers. The first stable-success budget identifies when the complete corrected state begins to remain correct at all later measured checkpoints. Each four-step training block provides one teaching example for the changed fact, so eight and 32 steps correspond to two and eight target-fact presentations, respectively. Budgets, exposure schedules, and outcome curves together provide a reproducible description of state updates.

\subsection{A Memory Interface for Physical Settings}
This update interface operates on scene facts. In factories, these may describe equipment configurations, process conditions, and operating rules; in warehouses and facility management, object attributes, zone conventions, and task conditions; and in everyday physical settings, environment configurations and collaboration requirements. Robots, production systems, and other Physical AI platforms can organize semantic-memory evaluation around the same requirement: revise a local fact while preserving the rest of the state.

Our experimental unit is a synthetic semantic configuration, and the measured outcome is its language-based readout after an update. Treating this memory layer as a separate component gives state writing, knowledge retrieval, and downstream task execution distinct interfaces. Joint correctness then characterizes changes between memory versions.

\section{Conclusion}
Motivated by changing knowledge in factories, warehouses, and everyday physical environments, we study parametric updates to the semantic-memory layer of Physical AI and present a joint evaluation of correction and preservation under finite update budgets. At fixed fact count and memory parameter count, the method combines budget trajectories, old/new answer diagnostics, and stable success at measured checkpoints to characterize how a complete state emerges after a local revision. In four-field synthetic configurations with fixed-A LoRA, increasing the budget from eight to 32 steps raises joint success from 6.25\% to 94.79\% and 91.15\% in the two primary histories, and removes exact obsolete-value outputs. The ranking reversal against the constant-content reference further illustrates why budget and preservation should be considered together. The complete protocol and measurements provide a reproducible component-level basis for evaluating memory updates for physical settings.

\section*{Author Contributions}
Jiaming Wu supervised the research.

\bibliographystyle{IEEEtran}
\bibliography{references}
\appendices

\section{Complete Budget Grid and Output Validity}
\label{app:grid}
\begin{figure*}[!tb]
\centering\includegraphics[width=\textwidth]{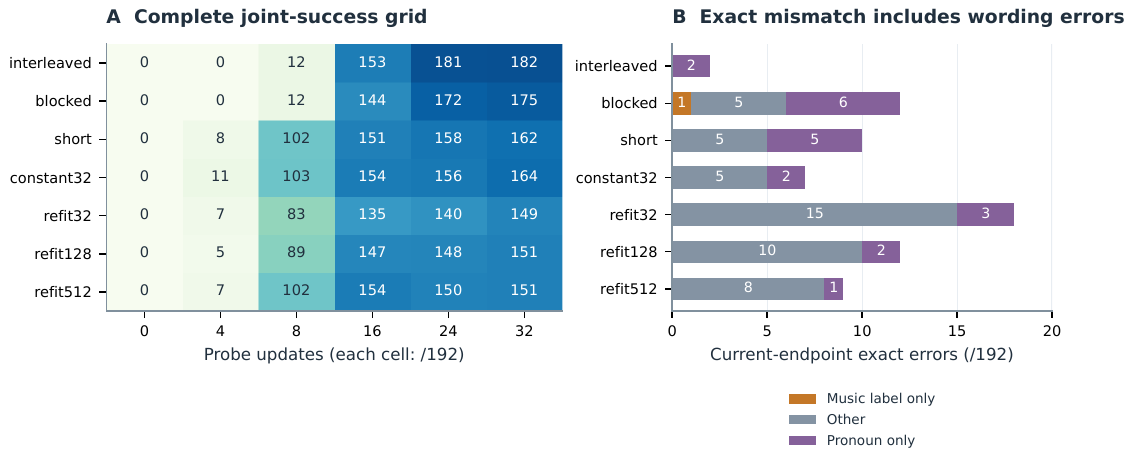}
\caption{Complete results. (A) Joint exact counts out of 192 at all predefined budgets, including every refit reference. (B) Current-endpoint exact mismatches partitioned by a string classifier into pronoun-only, music-label-only, and remaining lexical differences. The classification complements the original exact-match scores.}\label{fig:grid}
\end{figure*}
The full grid retains all seven branches and six budgets, including non-monotonic refit curves. At the current endpoint, Interleaved has two pronoun mismatches; Blocked has six pronoun mismatches, one music-label mismatch, and five longer non-target responses. The classifier distinguishes pronoun-only, music-label-only, and other mismatches. The categories describe lexical differences from the exact target sentence.

A separate generic probe contains 12 MCQs with four option rotations per question. The bare model scores 47/48. Mean current scores are 97.01\%/96.88\% for Interleaved/Blocked; final-probe scores are 97.14\%/96.91\%. The probe reports accuracy on these 12 questions, aggregating the four option rotations at each evaluated checkpoint.

\section{Implementation and Reproducibility}
The complete matrix contains eight initializations, 80 endpoints, and 448 single-topic probes: 536 completed jobs and 83,072 successful optimization steps. Each probe independently reloads its corresponding endpoint. Paired histories match the multiset of complete fact tables, teaching-row order, and anchor order. The complete fact table and its occurrence index determine the matched training seeds.

Independent reloads reproduce raw recall at steps 8 and 32. A CPU audit verifies 1,024 checkpoints (5,899,297,280 bytes), including hashes, metadata, finite tensors, frozen-A equality to paired initializations, and FP32-master casts to deployed BF16 values. Runtime parameter-version checks verify that base-model parameters remain unchanged. The primary order contrast is reconstructed from integer counts as exactly zero.

The accompanying material contains the fixed protocol, aggregate outcomes, likelihood margins, exact-output classifications, and the data used by each figure. The reported update budgets count successful optimization steps; loading, diagnostics, and independent evaluation are recorded as separate operations.

\end{document}